\pdfoutput=1

\documentclass[11pt]{article}

\usepackage{acl}

\usepackage{times}
\usepackage{latexsym}
\usepackage{hyperref}
\usepackage[T1]{fontenc}
 
\usepackage[utf8]{inputenc}

\usepackage{microtype}

\usepackage{inconsolata}

\usepackage{algorithmicx}
\usepackage{amsmath}
\usepackage{amsfonts}
\usepackage{amssymb}
\usepackage{enumitem}
\usepackage{multicol}
\usepackage{multirow}
\usepackage{array}
\usepackage{booktabs}
\usepackage{float}
\usepackage{graphicx}
\usepackage{amssymb} 
\usepackage{pifont}   
\usepackage{xcolor}   
\usepackage{subcaption}
\usepackage{hyperref}

\usepackage{amsmath,amsfonts,bm}

\def\eqref#1{equation~\ref{#1}}

\def\1{\bm{1}}

\DeclareMathAlphabet{\mathsfit}{\encodingdefault}{\sfdefault}{m}{sl}
\SetMathAlphabet{\mathsfit}{bold}{\encodingdefault}{\sfdefault}{bx}{n}

\def\gE{{\mathcal{E}}}

\def\gG{{\mathcal{G}}}

\def\gN{{\mathcal{N}}}

\def\gR{{\mathcal{R}}}

\def\sC{{\mathbb{C}}}

\def\sO{{\mathbb{O}}}

\def\sV{{\mathbb{V}}}

\usepackage[linesnumbered,ruled,vlined]{algorithm2e}

\newcommand\ee{$\textsc{EV}^{2}$}
\newcommand\ees{$\textsc{EV}^{2}$\ }

\newcommand{\gck}{{\color{green}\checkmark}} 
\newcommand{\rcs}{{\color{red}\ding{55}}} 

\definecolor{ForestGreen}{HTML}{009B55}
\definecolor{ProcessBlue}{HTML}{00B0F0}

\usepackage[figuresright]{rotating}
\usepackage[most]{tcolorbox}
\definecolor{BoxBackground}{RGB}{240, 240, 240} 
\definecolor{BoxFrame}{RGB}{0, 0, 0} 
\definecolor{TitleBackground}{RGB}{0, 0, 0} 
\definecolor{TitleText}{RGB}{255, 255, 255} 
\tcbset{
  academicbox/.style={
    boxsep=5pt,
    left=2pt,
    right=2pt,
    bottom=0.5pt,
    boxrule=0.5pt,
    colback=BoxBackground,
    colframe=BoxFrame,
    colbacktitle=TitleBackground,
    coltitle=TitleText,
    enhanced,
    attach boxed title to top left={yshift=-0.1in,xshift=0.1in},
    boxed title style={boxrule=0pt,colframe=white},
    title={#1},
  }
}

\newtcolorbox{AcademicBox}[1][]{academicbox=#1}
\definecolor{SoftBlue}{RGB}{135, 206, 250}  
\definecolor{SoftOrange}{RGB}{255, 224, 178} 
\definecolor{SoftGreen}{RGB}{144, 238, 144}  
\definecolor{CorrectGreen}{RGB}{76, 175, 80} 
\definecolor{ErrorRed}{RGB}{211, 47, 47} 

\title{A Comprehensive Evaluation on Event Reasoning of Large \\ Language Models}

\author{{\bf Zhengwei Tao$^{12}$ ~Zhi Jin$^{1}$\thanks{*Corresponding author.} ~Yifan Zhang$^{1}$ ~Xiancai Chen$^{1}$ ~Haiyan Zhao$^{1}$ ~Jia Li$^{1}$}\\{\bf ~Bing Liang$^{2}$ ~Chongyang Tao$^3$ ~Qun Liu$^4$ ~ Kam-Fai Wong$^2$} \\
        $^1$Key Lab of HCST (PKU), MOE; School of Computer Science, Peking University \\
        $^2$Department of Systems Engineering and Engineering Management, The Chinese University of Hong Kong \\
        $^3$Beihang University~$^4$Huawei Noah's Ark Lab\\ 
        \texttt{\{tttzw, xiancaich, yifanzhang\}@stu.pku.edu.cn}\\ ~\texttt{\{zhijin, zhhy.sei, lijiaa\}@pku.edu.cn},
        ~\texttt{chongyang@buaa.edu.cn}\\
        \texttt{bin.liang@cuhk.edu.hk}, ~\texttt{qun.liu@huawei.com}, ~\texttt{kfwong@se.cuhk.edu.hk}
        }

\begin{document}
\maketitle
\begin{abstract}
Event reasoning is a fundamental ability that underlies many applications. 
It requires event schema knowledge to perform global reasoning and needs to deal with the diversity of the inter-event relations and the reasoning paradigms. 
How well LLMs accomplish event reasoning on various relations and reasoning paradigms remains unknown. To mitigate this disparity, we comprehensively evaluate the abilities of event reasoning of LLMs. We introduce a novel benchmark \ees for EValuation of EVent reasoning. 
\ee consists of two levels of evaluation of schema and instance and is comprehensive in relations and reasoning paradigms. 
We conduct extensive experiments on \ee. 
We find that LLMs have abilities to accomplish event reasoning but their performances are far from satisfactory. 
We also notice the imbalance of event reasoning abilities in LLMs. 
Besides, LLMs have event schema knowledge, however, they're not aligned with humans on how to utilize the knowledge.
Based on these findings, we guide the LLMs in utilizing the event schema knowledge as memory leading to improvements on event reasoning. 
Code and Dataset are available on \href{https://github.com/TZWwww/EV2}{https://github.com/TZWwww/EV2}.

\end{abstract}

\section{Introduction}

\begin{figure}[!tb]
\setlength{\belowcaptionskip}{-5mm}
    \centering
    \includegraphics[width=1\columnwidth]{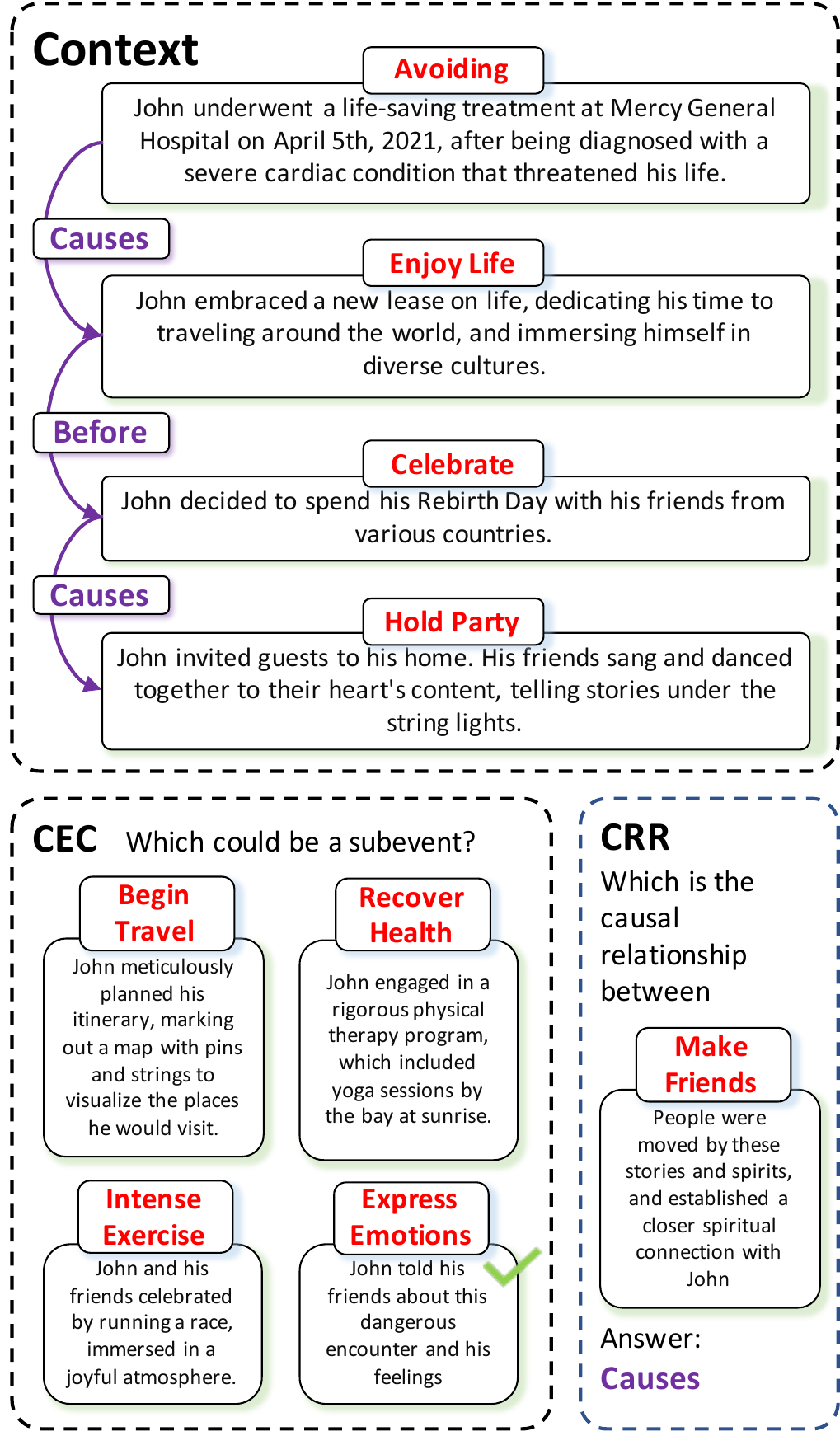}
    \caption{An example of event reasoning. The \textcolor{red}{red} words are event schema knowledge. The sentences below are event instances. In event reasoning, there are various paradigms such as Contextual Event Classification~(CEC) and Contextual Relation Reasoning~(CRR), and diverse inter-event relations.}
    \label{fig:intro}
\end{figure}

Events are instances or occurrences that form the basic semantic building units encompassing the meanings of Activities, Accomplishments, Achievements, and States~\cite{vendler1957verbs}. 
Event Reasoning is the ability to process and analyze events and their complex interconnections. 
Compared with other abilities, event reasoning is unique in some aspects. 
Firstly, it requires knowledge in the form of event schemas, capturing the progress of event evolution in scenarios, then performing global reasoning~\cite{li2021future, mao2021event}. 
As shown in Figure~\ref{fig:intro}, each event instance is associated with an event type. All event types and their relations form the event schema knowledge which reflects the logic and mechanism of event evolution. 
Knowing the event occurrence chain as "Avoiding", "Enjoy life", "Celebrate", and "Hold Party" would result in the following event "Express Emotions".
Second, the inter-event relations and reasoning paradigms are various. Event reasoning incorporates reasoning events according to a certain relation~\cite{du2022care, sap2019socialiqa} and reasoning inter-event relations~\cite{ning2018multi, caselli2017event}. The queried relations are diversified such as causality~\cite{roemmele2011choice}, temporality~\cite{zhou2019going}, and hierachy~\cite{glavavs2014hieve}. 
There are various paradigms such as reasoning the event or the inter-relation. 
As a fundamental competency within LLMs, event reasoning supports a multitude of Natural Language Processing (NLP) tasks, including recommendation engines~\cite{yang2020temporal}, interactive question-answer systems~\cite{souza2020event}, and AI Agents~\cite{liu2023agentbench}. 
Therefore, the enhancement of event reasoning abilities is essential for the advancement of LLMs.

LLMs like LLAMA~\cite{touvron2023llama} series and GPT series~\cite{brown2020language} have demonstrated exceptional accomplishments in various natural language reasoning~\cite{bang2023multitask,xu2023re}. 
Existing research has evaluated a broad spectrum of reasoning abilities of LLMs such as commonsence~\cite{bian2023chatgpt}, sentence relations~\cite{chan2023chatgpt}, and math~\cite{arora2023have}. 
However, studies on the comprehensive evaluation of event reasoning of LLMs are scarce. The incompleteness of current event reasoning evaluation is reflected in two aspects. First, current works only focus on instance-level events, resulting in unclearness of how LLMs understand and utilize the event schema knowledge~\cite{chan2023chatgpt}. Investigating the event knowledge of LLMs and how they employ them underlines applications such as event-based memory systems. Besides, they ignore the diversity of relations and paradigms~\cite{yuan2023zero}. Such findings could be biased since they neglect discrepancies brought by different aspects. These disparities hinge on the development of such crucial abilities of LLMs. 

In this paper, we comprehensively evaluate event reasoning in knowledge and abilities. Since there are existing datasets that are comprehensive in relations and paradigms, and can cover both levels of schema and instance, we introduce a novel benchmark \ees for the \textbf{EV}aluation of \textbf{EV}ent reasoning. \ees is featured in evaluating both aligned schema-level and instance-level. The schema-level evaluation investigates the event schema knowledge of LLMs while the instance-level testifies the event reasoning abilities. Besides, to evaluate event reasoning in various types of relation and reasoning paradigms, \ees includes two event reasoning tasks, namely Contextual Event Classification~(CEC) and Contextual Relation Reasoning~(CRR) as shown in Figure~\ref{fig:intro}. \ees is constructed from both GPT generation and human annotation. Utilizing \ee, we comprehensively evaluate how well LLMs do event reasoning in terms of abilities and knowledge. Specifically, we mainly explore four research questions: 
1) How proficient abilities of event reasoning do LLMs have?
2) To what extent do LLMs have the event schema knowledge?
3) Are LLMs aligned with humans in leveraging event schema knowledge?
4) Can LLMs perform better event reasoning with explicit guidance of leveraging event schema knowledge? 

We conduct extensive experiments on \ees to answer these questions. The results provide insights into event reasoning that: 
1) LLMs have the abilities of event reasoning, but are far from satisfactory and are imbalanced in different relations and reasoning paradigms. 
2) LLMs have event schema knowledge. They can answer the schema-level questions with similar accuracy to the instance-level questions. However, the development of schema-level abilities falls behind those of instance-level. 
3) LLMs are not aligned with humans in the aspect of leveraging event schema knowledge.
4) Based on the findings, we investigate guiding the LLMs to utilize event schema knowledge. With the guidance, LLMs can perform better event reasoning which sheds light on modeling event knowledge as memory of LLMs to enhance event reasoning.

We summarize our contributions as follows:
\begin{itemize}[topsep=0pt]
\setlength{\parskip}{0pt}
\setlength{\parsep}{-1pt}
\setlength{\leftmargin}{-1pt}
\item[$\bullet$] We evaluate event reasoning in both levels of schema and instance, and various relations and paradigms. 

\item[$\bullet$] We construct a novel benchmark \ees which features two levels of evaluation and comprehensive in relations and reasoning paradigms. We conduct extensive experiments to probe how LLMs perform event reasoning. 

\item[$\bullet$] We conclude several insights. Based on our findings, we guide LLMs to utilize event schema knowledge as memory achieving improvements in event reasoning.

\end{itemize}

\section{Problem Formulation}

Event reasoning is to anticipate the occurrences of certain relations or deduce interrelated correlations~\cite{tao2023eveval}.  
The relations encompass causality~\cite{du2022care}, temporality~\cite{zhou2019going}, and hierarchy~\cite{glavavs2014hieve}. 

Event reasoning requires comprehension of event schema knowledge. An event schema of a scenario is a schema-level graph $\gG^{s}=(\sV^{s}, \mathbb{E}^{s})$\footnote{Superscript $s$ represents schema level.}, where $\sV^{s}$ is the set of event types and $\mathbb{E}^{s}$ is the set of relations between events. Each edge in $\mathbb{E}^{s}$ is a relation triplet $(\gE_{i}^{s}, \gR, \gE_{j}^{s})$ standing for that there is the relation $\gR$ between $\gE_{i}^{s}$ and $\gE_{j}^{s}$. 
With instantiation, we have the instance-level event graph $\gG^{i}=(\sV^{i}, \mathbb{E}^{i})$\footnote{Superscript $i$ represents instance level.}. An instance event $\gE^{i}$ has an event type $\gE^{s}$ but with detailed event arguments and context~\cite{doddington2004automatic}. The nodes and edges of these two graphs are corresponding, namely, each triplet in $\gG^{s}$ has a corresponding triplet in $\gG^{i}$ with the same inter-relation. In both levels, we consider totally six relation types, namely $\gR\in$\{\texttt{Causes}, \texttt{IsResult}, \texttt{Before}, \texttt{After}, \texttt{IsSubevent}, \texttt{HasSubevent}\}\footnote{In this work, we denote the direction of the relation to that the former event have relation to the latter event. For example, ($\gE_i$, \texttt{Causes}, $\gE_j$) stands for $\gE_i$ causes $\gE_j$.}. \texttt{Causes} and \texttt{IsResult} are Causal relations, \texttt{Before} and \texttt{After} belong to Temporal type while \texttt{IsSubevent} and \texttt{HasSubevent} are Hierarchical type.

\ees consists of two event reasoning paradigms for both levels of schema and instance. The first is Contextual Event Classification~(CEC) and the second is Contextualized Relation Reasoning~(CRR). 

\paragraph{CEC} Given graph $\gG$, either schema- or instance-level, queried event $\gE\in\gG$, and target relation $\gR$, CEC requires the model to answer an event $\gE_{a}$:
\begin{equation}
\label{cec}
    \gE_{a} = \mathrm{M}(\gE, \gR, \gG, \sC).
\end{equation}
$\mathrm{M}$ is the model, $\sC$ is the candidate event set. CEC evaluates the model's comprehension of event semantics and structure. 

\paragraph{CRR} Given graph $\gG$, either schema- or instance-level, two queried events $\gE_i, \gE_j\in\gG$, CRR requires to determine the relation $\gR$ between them:
\begin{equation}
\label{crr}
    \gR = \mathrm{M}(\gE_i, \gE_j, \gG).
\end{equation}
CRR evaluates the understanding of event relations.

In both schema and instance levels, \ees has CEC and CRR tasks. Schema-level tasks require models to be rich in knowledge while tasks for instance need models to process detailed information.

\section{Benchmark Construction}
Constructing the \ee~benchmark is challenging since events and their relations are semantically abstract concepts compared with entity concepts. The occurrence of events and their relations not only follow objective natural laws but are also influenced by social and humanistic factors. Therefore, annotating such data is extremely label-intensive leading to a lack of evaluation dataset of the task. Previous works mainly construct such datasets by extracting events and relations from some unlabeled corpus such as news reports~\cite{caselli2017event, ning2018multi, ogorman-etal-2016-richer}. However, such a method suffers from limited event relational patterns and domain-specific language expression. To mitigate such problems, in this work, we construct our comprehensive event reasoning evaluation from scratch.

In \ee~, we evaluate event reasoning in various relation and reasoning paradigms of both schema and instance levels. In our pilot trial, we directly annotate each question and answer by human annotators. We find this process extremely time-consuming since human annotators must imagine scenarios and guarantee their correctness. Besides, the diversity of the events and their relations are poor. Annotators only write the most common events. Therefore, we propose a four-stage construction process. Generally, we first synthesize both the schema and corresponding instance event prompting-graphs automatically as prompts for later annotations. Due to the large size of the synthesis graphs, before formal annotating, we then recruit annotators to remove incorrect graphs which are hard to modify. After that, the annotators curate schema and instance graphs based on the generated prompts. Finally, we adapt the graphs into questions and answers. We celebrate this process in the following sections.

\subsection{Prompting-Graph Synthesizing}
To cover more domain and granular scenarios, we synthesize event graphs as prompts for annotations instead of annotating from scratch. Specifically, we first establish the schema graph $\gG^{s}$. Then we employ GPT4 to generate the instance graph $\gG^{i}$. 

\paragraph{Schema Graph} The schema graph represents event occurrence knowledge. Harvesting such knowledge is a research point~\cite{du-etal-2022-resin}. We here leverage EECKG~\cite{wang2022ecckg} to ensure a diverse range of event types in our schema. EECKG combines rule-based reasoning with crowdsourced insights, built on ConceptNet's structure. Nodes in EECKG represent verb phrases as events, and edges denote inter-event relations, focusing on \texttt{Causes}, \texttt{Before}, and \texttt{HasSubevent}.

Our objective mandates that the nodes within $\gG^{s}$ should represent event types. Therefore, we filter EECKG nodes, removing concrete event instances. Preference is given to nodes with at most two words, as longer descriptions tend to include specific details.
For events with fewer than two words, we use GPT4 to enhance our selection, ensuring the appropriate abstraction level for our schema graph with the following prompt:


\begin{AcademicBox}
    \small
    \vspace{-2mm}
    \textbf{\#\#\# Instructions:} \\
    Determine which of the following candidate phrases are abstract and conceptual event types.
\end{AcademicBox}

We identify a subset of remaining events that are too generic. To refine the event selection, we also exclude the most frequent events from our subset to avoid generic events. 

We then dissect the interconnected EECKG into separate components, each representing a distinct scenario. To prevent semantic drift, we carefully control the size of each component. Starting from a node, we conduct a random walk until the number of nodes surpasses a set threshold, thus defining a component. This process is executed for all nodes to gather all components, as detailed in Algorithm~\ref{algorithm}. Post-extraction, we eliminate cycles to convert these structures into DAGs. 

EECKG only contains forward event evolution relations such as \texttt{Causes}. We further include components of backward relations. We generate a reversed version for each component by inverting edge directions and replacing relations with their opposites: \texttt{IsResult}, \texttt{After}, and \texttt{IsSubevent}. This creates the backward components.

\begin{algorithm}
\caption{Components Construction}
\label{algorithm}
\SetKwInOut{Input}{Input}
\SetKwInOut{Output}{Output}
\SetKwFunction{RandomWalk}{RandomWalk}
\SetKwFunction{RandomInt}{RandomInt}
\SetKwProg{Fn}{Function}{:}{}
\SetKwFunction{Append}{Append}
\SetKwFunction{Sample}{Sample}
\SetKwFunction{Neighbors}{Neighbors}

\Input{EECKG $\gG$, $\gN$}
\Output{A list of components $\sO$.}

$\sO$ = [\ ] \\

\Fn{\RandomWalk{$start$, $c$}}{
     $l$ = \RandomInt{$\gN$, $\gN$+2}\\
    \If {$l \leq \text{len}(c)$}{
        $c$.\Append($start$)\\
        \KwRet $c$
    }
    $n$ = \Sample{start.\Neighbors}\\
    \If {$n \notin c$}{
        $c \gets \RandomWalk(n, c \cup \{n\})$ \\
        \KwRet $c$

    }
    
    \KwRet \text{null} 
}

\ForEach{$node \in G$}{
     $component$ = \RandomWalk($node$, [\ ])\\
     $\sO$.\Append($component$)
}

\KwRet $\sO$

\end{algorithm}

In preparation for constructing tasks for CEC and CRR, we label two events for each component. We sample three event pairs $(\gE_h, \gE_t)$ per component with a maximum inter-path length of four, utilizing their predecessors as background events. These pairs and background events form a schema graph. When the path length between $\gE_h$ and $\gE_t$ is two, the direct relation serves as the queried relation; for longer paths, we deduce the relation using Table~\ref{rule}. We construct a schema graph, queried event pair, and their relation $(\gE_h, \gE_t, \gR, \gG^{s})$.

\paragraph{Instance Graph} We next harvest instance graph $\gG^{i}$ for each schema graph $\gG^{s}$. For each node $\gE^{s}\in\gG^{s}$, we ask GPT4 to generate $\gE^{i}$ using the following prompt:

\begin{AcademicBox}
    \small
    \vspace{-2mm}
    \textbf{\#\#\# Instructions:} \\
    Generate an instance event for each abstract event. The abstract event is the event type of the instance event. All the instance events form a coherent story which maintain the relations of each abstract event. The integrated story should have explicit roles, location, and time. The whole story should be detailed, diverse in topic and scenarios, and rich in knowledge.
\end{AcademicBox}


We inherit the relations of $\gG^{s}$ and obtain $\gG^{i}$. We naturally obtain the instances of $\gE_h$ and $\gE_t$. Finally, we obtain 1,600 schema prompting graphs and 1,600 corresponding instance graphs.

\subsection{Manual Filtering}
After curating the prompting graphs of both schema and instance levels, the next is to annotate based on the prompting graphs. However, we find some of the prompting graphs are incorrect and hard to modify. Therefore, before formal annotation, we launch another manual filtering step to remove such graphs.

We then recruit 8 well-educated human annotators where they each process 200 data. Their missions are to investigate the $\gG^{s}$ and $\gG^{i}$ by the following steps:

\begin{itemize}[topsep=0pt]
    \setlength{\itemsep}{0pt}
    \setlength{\parskip}{0pt}
\setlength{\leftmargin}{-1pt}
\item[1)] Check whether $\gG^{s}$ can be modified, the events are abstract, the relations are correct, and $\gG^{s}$ tells an entire story of a scenario.

\item[2)] Check whether $\gG^{i}$ can be modified., the events are concrete, the relations are correct, and $\gG^{i}$ tells an entire story of a scenario.

\item[3)] Check whether $\gG^{s}$ and $\gG^{i}$ are consistent, there are obvious schema-instance relations between them. 

\end{itemize}

If any of these conditions are not met, we discard this datum. After this filtering process, we remain 491 graph pairs.

\begin{table}[!t]
\centering
\footnotesize
\setlength{\tabcolsep}{2.7mm}{\begin{tabular}{lc}

\toprule
\toprule
\textsc{Rule}&\textsc{Induction}\\
\midrule
(\texttt{Before})$^+$&\texttt{Before}\\
(\texttt{After})$^+$&\texttt{After}\\

(\texttt{Before})$^{\star}$(\texttt{Causes})$^+$(\texttt{Before})$^{\star}$&\texttt{Before}\\

(\texttt{After})$^{\star}$(\texttt{IsResult})$^+$(\texttt{After})$^{\star}$&\texttt{After}\\

(\texttt{Before})$^{\star}$(\texttt{HasSubevent})$^+$( \texttt{Before})$^{\star}$&\texttt{Before}\\
(\texttt{Causes})$^{\star}$(\texttt{HasSubevent})$^+$( \texttt{Causes})$^{\star}$&\texttt{Causes}\\

(\texttt{After})$^{\star}$(\texttt{IsSubevent})$^+$( \texttt{After})$^{\star}$&\texttt{After}\\
(\texttt{IsResult})$^{\star}$(\texttt{IsSubevent})$^+$( \texttt{IsResult})$^{\star}$&\texttt{IsResult}\\

\bottomrule

\end{tabular}}
\caption{Relation induction rules. $^{\star}$ denotes there exists zero or more. $^+$ means there is at least one.}
\label{rule}
\end{table}

\begin{table}[!t]
\centering
\footnotesize
\setlength{\tabcolsep}{1.5mm}{\begin{tabular}{cccccc}

\toprule
\toprule
\textsc{S-CEC}&\textsc{I-CEC}&\textsc{S-CRR}&\textsc{I-CRR}&\textsc{Graph Pairs}\\
\midrule
 492 & 491 & 730 & 735 & 491 \\

\bottomrule
\end{tabular}}
\caption{Number of \ee. \textsc{S} and \textsc{I} are schema and instance.}
\label{stats}
\end{table}

\subsection{Annotation}
In this stage, we formally annotate based on filtered prompting-graphs. The missions of this stage are to 1) rewrite the events and relations in both schema and instance graphs to make them strictly valid. 2) identify a query event $\gE_{h}^{s}$ and an answer event $\gE_{t}^{s}$ in the graph for later question adaptation. 3) write candidates as negative choices considering the answer event where each candidate event consists of a schema and an instance event. For the second mission, regarding schema and instance head events as the query and the tail as an answer, we ask GPT4 to generate 15 possible candidate instance events with their event types for prompting. 

We recruit another 10 annotators. The annotation is completed on our annotating system. Each annotator should rewrite correct data alongside at least the following standards:

\begin{itemize}[topsep=0pt]
    \setlength{\itemsep}{0pt}
    \setlength{\parskip}{0pt}
\setlength{\leftmargin}{-1pt}
\item[1)] Rewrite $\gG^{s}$ and $\gG^{i}$ making them correct as high-probability knowledge, and do not consider low-probability situations. 

\item[2)] Rewrite $\gG^{s}$ and $\gG^{i}$ leading to the coherence of the whole graph, and there's no semantic drift.

\item[3)] Rewrite $\gG^{s}$ and $\gG^{i}$ making them consistent. Schema events and instance events require a clear distinction in hyper-hypo relation.

\item[4)] Rewrite the target event making it only can be answered when considering the whole graph.

\item[5)] Rewrite the instance events making them should be expressed independently without connective expressions such as "After $\gE_1$". This could incur information leakage.

\end{itemize}

We describe the detailed annotation process in the Appendix.

\subsection{Question Adaptation}
The last step is to construct questions of CEC and CRR in both schema and instance levels based on the annotated graphs. 
We use the schema part of annotation as the schema-level questions and the instance part as instance-level questions. 

For schema-level CEC, we naturally use the queried event $\gE_{h}^{s}$ and other nodes except for the answer event $\gE_{t}^{s}$ as context to form a question. Then we use the answer event $\gE_{t}^{s}$ as the answer. We do similarly at the instance level.
For CRR, we regard $\gE_{h}^{s}$ and $\gE_{t}^{s}$ as queried events and use the relation between them as the answer to form the schema-level question. For instance part, we adopt a similar way. 

Finally, our CEC task is a 4-way multiple-choice task. The CRR is a 3-way multiple-choice task. In CRR, the choices for temporal, causal, and hierarchy relations are [\texttt{Before}, \texttt{After}, \texttt{Vague}], [\texttt{Causes}, \texttt{IsResult}, \texttt{None}], and [\texttt{IsSubevent}, \texttt{HasSubevent}, \texttt{None}] respectively. We show examples of both tasks in Figure~\ref{fig:intro}. 
We report the number of each task and the average nodes and edges of \ee~in Figure~\ref{stats}. 

\subsection{Quality Inspection}
After the annotation, to further guarantee the validation of \ee, we recruit another group of human annotators to inspect the quality of \ee. We sample 100 data for all tasks. We ask them to give two scores for each sample: 

\begin{itemize}[leftmargin=1mm, itemindent=0.2cm]
    \setlength{\itemsep}{0pt}
    \setlength{\parskip}{0pt}
    \item[] \textit{Correct:} Rate 1 if correct, otherwise rate 0.
    \item[] \textit{Contextualized:} Rate 1 if the answer relies on the context events, otherwise rate 0.
\end{itemize}

Finally, we get 91\% for \textit{Correct} and 92\% for \textit{Contextualized}. Human examination testifies that \ees is qualified. Besides, context events count.

\subsection{Existing Dataset Comparison}
We compare our benchmark to existing related datasets. We show detailed comparison in Table~\ref{comparison}. Our benchmark is the only one that is for contextualized event reasoning of various relations and paradigms on both schema and instance levels. 

\begin{table}[!t]
\centering
\footnotesize
\setlength{\tabcolsep}{1.5mm}{\begin{tabular}{lcccc}

\toprule
\toprule
\textsc{Dataset}&\textsc{L}&\textsc{C}&\textsc{M-R}&\textsc{M-P}\\
\midrule
ALTLEX\cite{hidey-mckeown-2016-identifying} & \textit{I} & \rcs & \rcs & \rcs \\
ASER\cite{zhang2020aser} & \textit{S} & \rcs & \gck & \rcs \\
ATOMIC\cite{sap2019atomic} & \textit{S} & \rcs & \gck & \rcs \\
COPA\cite{roemmele2011choice} & \textit{I} & \rcs & \rcs & \rcs \\
CQA\cite{bondarenko-etal-2022-causalqa} & \textit{I} & \gck & \gck & \rcs \\
ECARE\cite{du2022care} & \textit{I} & \rcs & \rcs & \rcs \\
ESL\cite{caselli2017event} & \textit{I} & \gck & \rcs & \rcs \\
ESTER\cite{han2021ester} & \textit{I} & \gck & \gck & \rcs \\
HIEVE\cite{glavavs2014hieve} & \textit{I} & \gck & \rcs & \rcs \\
KAIROS\cite{li2021future} & \textit{S} & \gck & \rcs & \rcs \\
LDC2020E25\cite{li2021future} & \textit{S} & \gck & \rcs & \rcs \\
MATRES\cite{ning2018multi} & \textit{I} & \gck & \rcs & \rcs \\
MAVEN-ERE\cite{wang2022maven} & \textit{I} & \gck & \gck & \rcs \\
MCNC\cite{MarkGW-AAAI16} & \textit{I} & \gck & \rcs & \rcs \\
MCTACO\cite{zhou2019going} & \textit{I} & \gck & \rcs & \rcs \\
RED\cite{ogorman-etal-2016-richer} & \textit{I} & \gck & \gck & \rcs \\
SCITE\cite{li2021causality} & \textit{I} & \gck & \rcs & \rcs \\
SCT\cite{mostafazadeh2016corpus} & \textit{I} & \gck & \rcs & \rcs \\
SocialIQA\cite{sap2019socialiqa} & \textit{I} & \gck & \gck & \rcs \\
TB-Dense\cite{cassidy2014annotation} & \textit{I} & \gck & \rcs & \rcs \\
TRACIE\cite{zhou2020temporal} & \textit{I} & \gck & \rcs & \rcs \\
\midrule
\ee & \textit{S} \textit{I} & \gck & \gck & \gck \\

\bottomrule

\end{tabular}}
\caption{Comparison with existing event reasoning datasets. \textsc{L} stands for the included levels. \textsc{C} represents whether it's contextualized. \textsc{M-R} and \textsc{M-P} means if it has multi-relations and paradigms. \textit{S} and \textit{I} stand for schema and instance level.}
\label{comparison}
\end{table}

\section{Experiments}

\begin{figure*}[!tb]
    \centering
    \includegraphics[width=2.1\columnwidth]{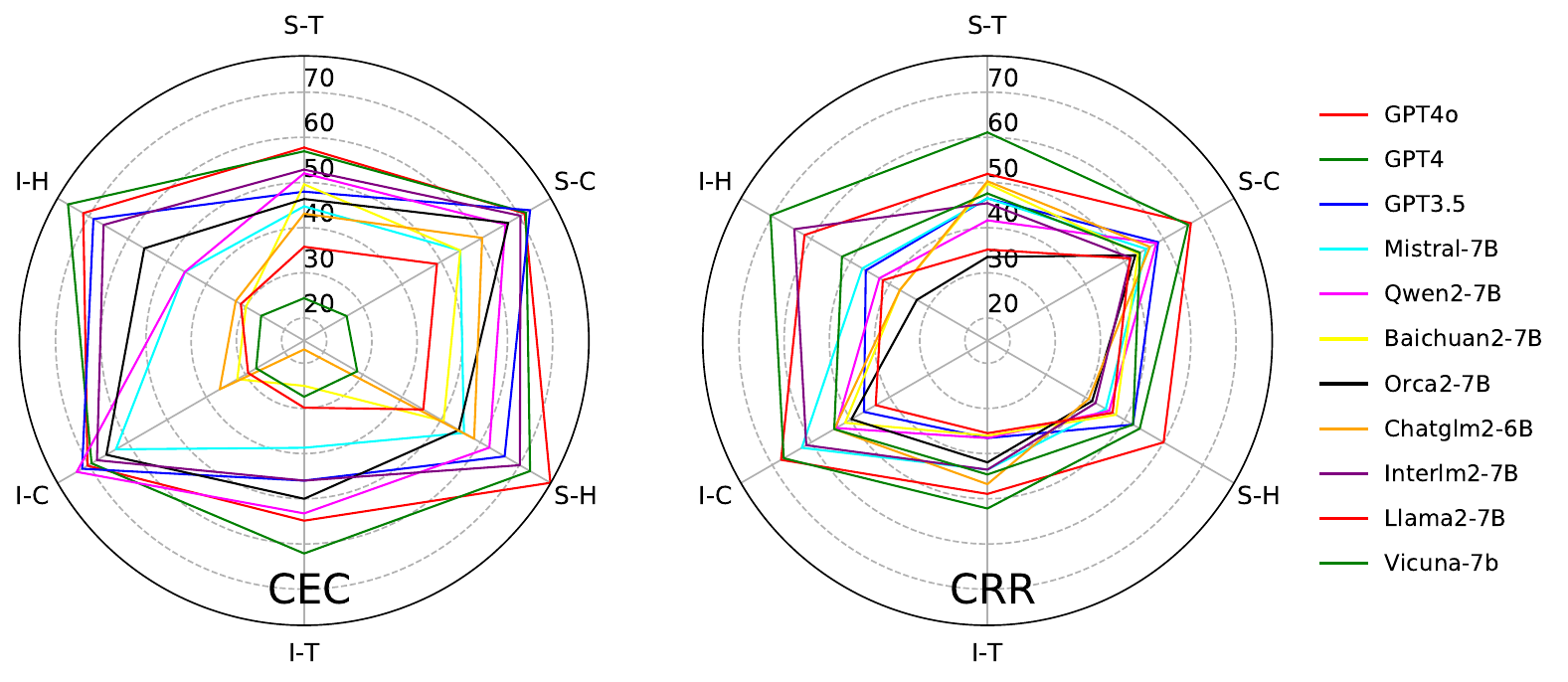}
    \caption{Results of CEC and CRR. S and I stand for schema- and instance-level. Relation types of Causality, Temporality, and Hierarchy are denoted as C, T, and H.}
    \label{fig:main}
\end{figure*}

\subsection{Evaluated LLMs}
We evaluate 11 LLMs on event reasoning. For the open-source models, we evaluate their chat-version. We evaluate GPT4o, GPT4, GPT3.5. For the closed-source models, we utilize their official APIs to conduct performance evaluations. For the open-source models, we include Mistral-7B~\cite{jiang2023mistral}, Qwen2-7B~\cite{yang2024qwen2}, Baichuan-2-7B~\cite{yang2023baichuan}, Orca2-7B~\cite{mitra2023orca}, Chatglm2-6B~\cite{glm2024chatglm}, Internlm2-7B~\cite{cai2024internlm2}, Llama2-7B~\cite{touvron2023llama}, and Vicuna-7B~\cite{vicuna2023}. Without loss of generosity, we use the model names to refer to the chat versions in the rest of our paper. For all evaluated LLMs, we use the same prompt. We show all prompts in the Appendix.

\section{Results and Findings}







\subsection{How proficient abilities of event reasoning do LLMs have?}

In this part, we mainly probe the abilities of how existing LLMs complete the event reasoning of the instance level.

\begin{table}
\centering
\small
\setlength\tabcolsep{7pt}
\begin{tabular}{lcccccccc}
\toprule
\toprule
Model & S-CEC & I-CEC & S-CRR & I-CRR \\
\midrule
GPT4o & 68.93 & 66.60 & 62.05 & 62.04 \\
GPT4 & 68.11 & 68.43 & 63.01 & 63.81 \\
GPT3.5 & 65.43 & 64.77 & 54.52 & 43.95 \\
Mistral-7B & 52.47 & 54.18 & 51.64 & 55.24 \\
Qwen2-7B  & 62.14 & 63.75 & 52.05 & 47.62 \\
Baichuan2-7B & 52.88 & 29.94 & 51.64 & 45.31 \\
Orca2-7B  & 59.88 & 60.08 & 46.16 & 45.17 \\
Chatglm2-6B  & 55.76 & 30.96 & 52.47 & 49.66 \\
Interlm2-7B  & 65.84 & 62.12 & 48.63 & 57.28 \\
Llama2-7B  & 45.06 & 29.74 & 46.58 & 41.22 \\
Vicuna-7b & 25.93 & 27.09 & 52.05 & 51.43 \\
\bottomrule
\end{tabular}
\caption{Average performances with updated models. S and I stand for schema- and instance-level. }
\label{tab:avg}
\end{table}

\paragraph{LLMs have the abilities of event reasoning, but even the strongest GPT-4 is far from satisfactory.} We evaluate CEC and CRR at the instance level. We show the results of different relations in Figure~\ref{fig:main} and detailed results in the Appendix. For CEC, GPT4 performs the best. 
Among all open-source LLMs, Qwen2 is the best while Internlm2 holds the second.
Early models such as Baichuan2-7B, Chatglm2-6B, Llama2-7B, and Vicuna-7B fall in this task where they score under 40\%.
For CRR, GPT4 excels all other models as well. Among all open-source LLMs, Internlm2-7B and Mistral-7B performs in the first tie. There is no obvious difference in the performance of other models for CRR. 

We show the average performance of instance-level CEC and CRR in columns I-CEC and I-CRR in Table~\ref{tab:avg}. Overall, existing LLMs such as GPT4, and Qwen2-7B have CEC event reasoning abilities. However, even the strongest GPT4 can only achieve 68.43 (4-way multiple choice) and 63.81 (3-way multiple choice) accuracy in each task showing there's much room for improvements of event reasoning.

\paragraph{The abilities of LLMs to deal with different relations and reasoning paradigms are unbalanced.}
Comparing CEC to CRR, as relation-wise results shown in Figure~\ref{fig:main} and average performances in columns I-CEC and I-CRR in Table~\ref{tab:avg}, LLMs perform better for CEC than CRR (note that CEC is a 4-way multiple choice task while CRR is of 3-way). 
To compare, we compute the average scores of I-CEC and I-CRR on models achieving above 40\% and 30\%\footnote{Models under these scores may lack statistic significance.}. 
We find I-CEC is much higher than I-CRR, with average scores 62.84 and 53.58. The results significantly suggest that CRR is harder and insifeciently solved than CEC. Existing pretraining and SFT datasets may be biased in paradigms. 

We then analyze performances on different relations. We compute the average scores of relations on models achieving above 40\% and 30\% on average I-CEC and I-CRR. The I-CEC average scores of Temporal, Causal, Hierarchical are 50.11,	68.71, and 61.22 while in I-CRR the scores are 43.31, 58.36, and 52.17. With these results alongside scores shown in Figure~\ref{fig:main}, LLMs perform best in Causal relation. Then, Temporal relation is the worst. It indicates current training can enable LLMs to reason causality. It also trains the event hierarchy comprehension. However, temporal reasoning is the hardest. More methods should be established to handle this problem. That further indicates the imbalance training of different relations. Methods and datasets of balanced abilities on relations are needed. Transferring abilities of different relations could also be feasible~\cite{tao2023unievent}.

This is a crucial finding. \citet{chan2023chatgpt} conduct causal event classification such as ECARE~\cite{du2022care}, and relation reasoning such as MATRES~\cite{ning2018multi}. They directly compare these two groups of results and conclude the gaps are merely from differences in relations. However, they ignore the difference in reasoning paradigms. Leveraging \ee, with disentangling relations and formulations, we investigate event reasoning with less bias.

\begin{figure}[!tb]
\setlength{\belowcaptionskip}{-5mm}
\setlength{\abovecaptionskip}{-1mm}
    \centering
    \includegraphics[width=1\columnwidth]{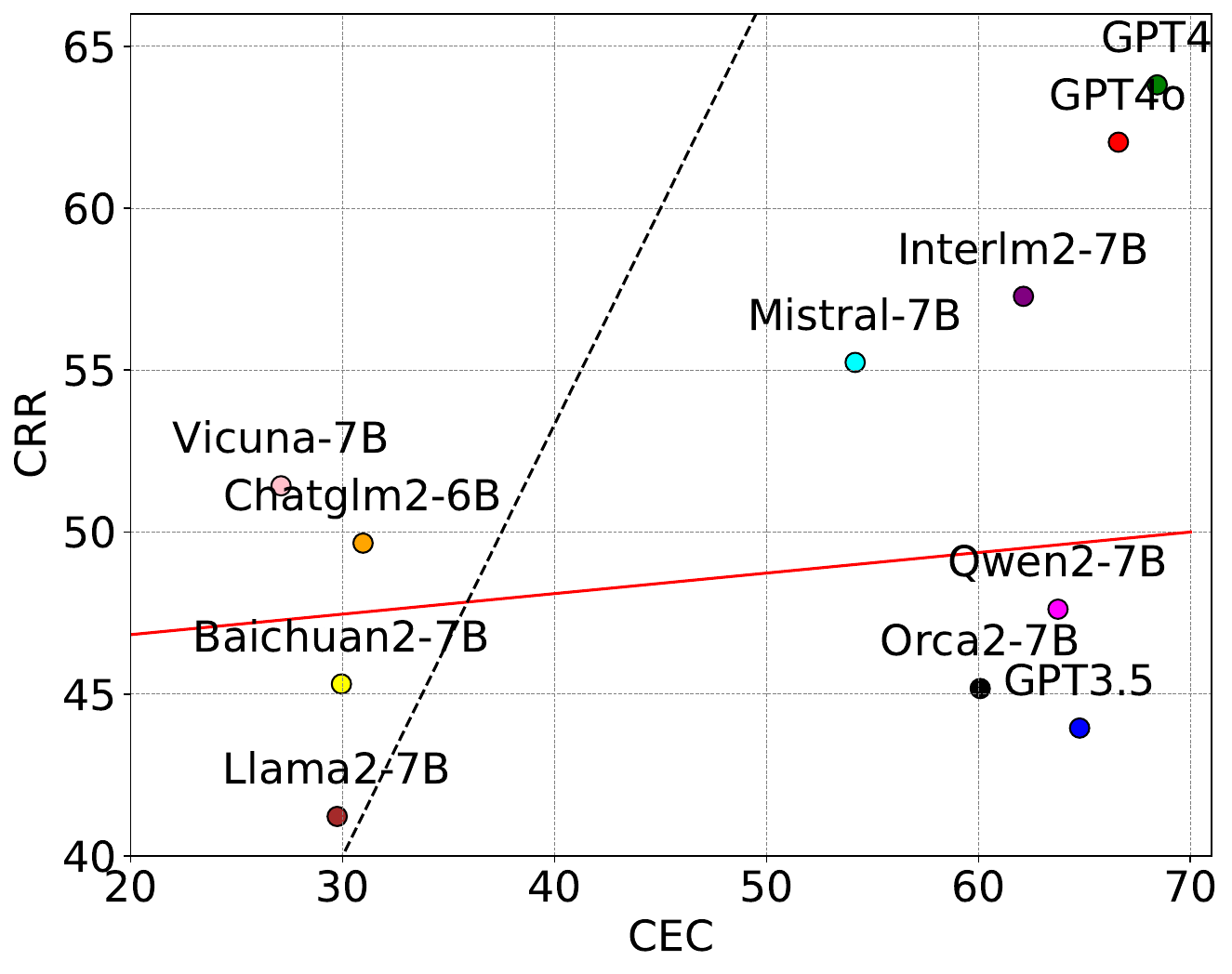}
    \caption{Improvements trend on CEC and CRR. The dashline represents the balanced improvement with slope 3/4 considering the CEC is a 4-way multiple-choice task while CRR has three choices. The \textcolor{red}{red} line is the regression line of models except GPT4.}
    \label{fig:cec-crr}
\end{figure}

\paragraph{CEC improves faster than CRR with model development.}
We investigate the improvement trends of CEC and CRR. In Figure~\ref{fig:cec-crr}. When models have poor event reasoning abilities, their CEC performances lie around the balanced line showing no significant differences in tasks. With the development, the CEC improves much faster than CRR for all models. This investigation further appeals to the need for training in comprehensive and balanced event reasoning abilities.

\begin{figure*}
    \centering
    \includegraphics[width=2\columnwidth]{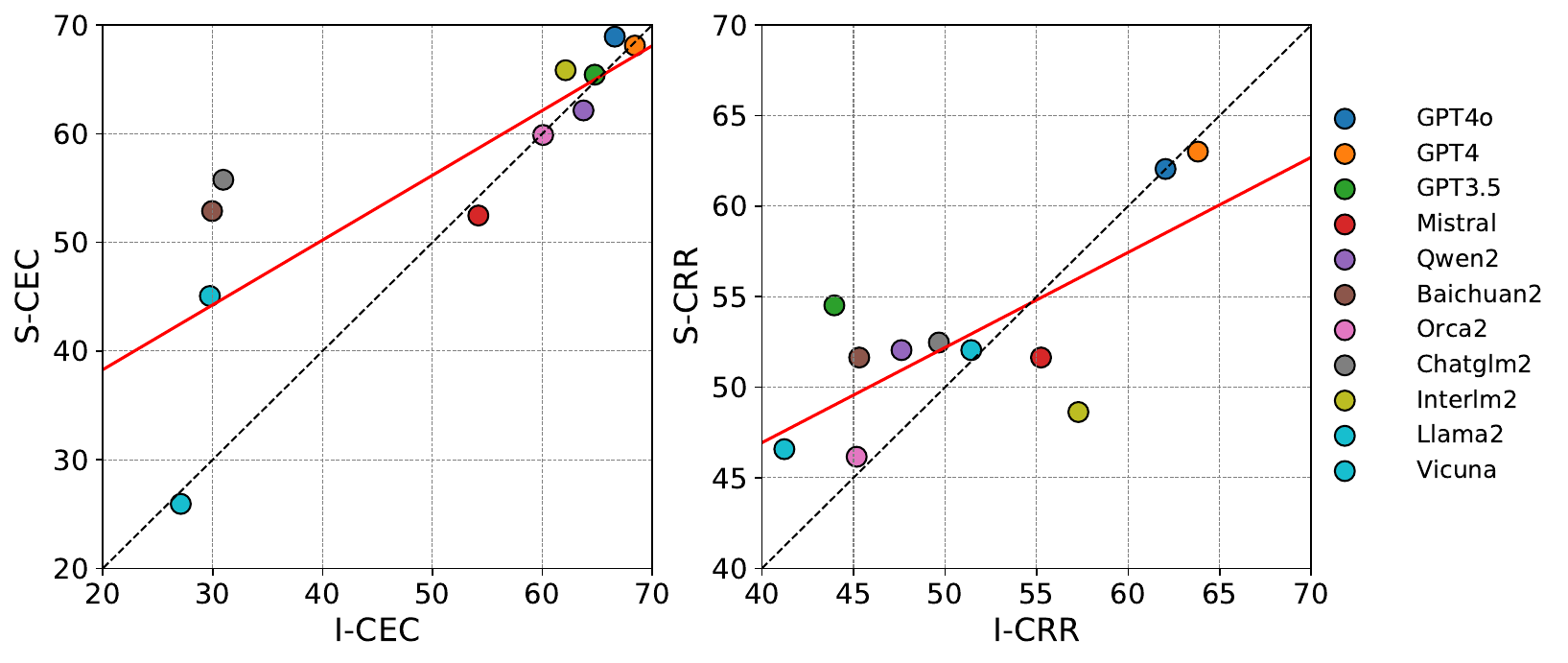}
    \caption{Comparisons between performances on instance- and schema-level. The dashed line represents the balanced improvement with slope 1. The \textcolor{red}{red} line is the regression line of all models}
    \label{fig:S-I}
\end{figure*}

\begin{table}
\centering
\small
\begin{tabular}{lcccc}
\toprule
\toprule
& \multicolumn{2}{c}{\textsc{CEC}} & \multicolumn{2}{c}{\textsc{CRR}} \\
\cmidrule{2-3}
\cmidrule{4-5}
 & \textsc{ET} & \textsc{Rel} & \textsc{ET} & \textsc{Rel}  \\
\midrule
GPT4o	& 65.53 & 40.23	& 69.84 & 51.24 \\
GPT4	&72.78 & 40.57 & 73.19 & 52.79 \\
GPT3.5	& 10.05 & 15.58 & 16.19 & 28.63 \\
Mistral-7B & 22.93 & 16.28 & 25.00 & 23.87 \\
Qwen2-7B & 11.40 & 15.97 & 13.07 & 16.58 \\
\bottomrule
\end{tabular}
\caption{Event schema knowledge Alignment. \textsc{ET} is the event type accuracy. \textsc{Rel} is relation triplet F1-score.}
\label{tab:align}
\end{table}

\begin{table*}
    \setlength{\belowcaptionskip}{-3mm}
    \centering
    \small
    \setlength\tabcolsep{12pt}
    \begin{tabular}{lcccccc}
        \toprule
        Model & Temporal & Causal & Hierarchical & \textsc{w.t.S} & \textsc{w.o.S} & $\Delta$ \\
        \midrule
        GPT4o & 58.06 & 73.45 & 85.71 & 71.49 & 66.60 & \textcolor{ForestGreen}{4.89$\uparrow$} \\ 
        GPT4 & 61.29 & 75.86 & 77.92 & 72.51 & 68.43 & \textcolor{ForestGreen}{4.08$\uparrow$} \\ 
        GPT3.5 & 52.42 & 77.59 & 77.92 & 71.28 & 64.77 & \textcolor{ForestGreen}{6.51$\uparrow$} \\ 
        Mistral-7B & 45.16 & 70.34 & 74.03 & 64.56 & 54.18 & \textcolor{ForestGreen}{10.38$\uparrow$} \\ 
        Qwen2-7B & 61.29 & 78.97 & 77.92 & 74.34 & 63.75 & \textcolor{ForestGreen}{10.59$\uparrow$} \\ 
        Baichuan2-7B & 41.13 & 56.90 & 57.14 & 52.95 & 29.94 & \textcolor{ForestGreen}{23.01$\uparrow$} \\ 
        Orca2-7B & 53.23 & 77.59 & 76.62 & 71.28 & 60.08 & \textcolor{ForestGreen}{11.20$\uparrow$} \\ 
        Chatglm2-6B & 38.71 & 65.86 & 54.55 & 57.23 & 30.96 & \textcolor{ForestGreen}{26.27$\uparrow$} \\ 
        Interlm2-7B & 57.26 & 80.69 & 81.82 & 74.95 & 62.12 & \textcolor{ForestGreen}{12.83$\uparrow$} \\ 
        Llama2-7B & 36.29 & 45.86 & 49.35 & 43.99 & 29.74 & \textcolor{ForestGreen}{14.25$\uparrow$} \\ 
        Vicuna-7b & 27.42 & 28.62 & 27.27 & 28.11 & 27.09 & \textcolor{ForestGreen}{1.02$\uparrow$} \\ 
        \bottomrule
    \end{tabular}
    \caption{Direct guidance with schema knowledge on instance level CEC task. \textsc{w.t.S} and \textsc{w.o.S} stands for average performances with and without event knowledge guidance. $\Delta$ is the difference between them.}
    \label{tab:cec direct}
\end{table*}

\begin{table*}
    \setlength{\belowcaptionskip}{-3mm}
    \centering
    \small
    \setlength\tabcolsep{12pt}
    \begin{tabular}{lcccccc}
        \toprule
        Model & Temporal & Causal & Hierarchical & \textsc{w.t.S} & \textsc{w.o.S} & $\Delta$ \\
        \midrule
        GPT4o          & 56.45    & 73.04  & 75.65     & 69.25        & 62.04 & \textcolor{ForestGreen}{7.21$\uparrow$} \\
        GPT4           & 53.23    & 70.74  & 68.70      & 65.99        & 63.81 & \textcolor{ForestGreen}{2.18$\uparrow$} \\
        GPT3.5         & 44.09    & 59.22  & 47.83     & 53.61        & 43.95 & \textcolor{ForestGreen}{9.66$\uparrow$} \\
        Mistral-7B     & 48.92    & 63.36  & 50.43     & 57.69        & 55.24 & \textcolor{ForestGreen}{2.45$\uparrow$} \\
        Qwen2-7B       & 47.31    & 66.13  & 55.65     & 59.73        & 47.62 & \textcolor{ForestGreen}{12.11$\uparrow$} \\
        Baichuan2-7B   & 46.24    & 58.29  & 40.87     & 52.52        & 45.31 & \textcolor{ForestGreen}{7.21$\uparrow$} \\
        Orca2-7B       & 44.09    & 65.9   & 51.30      & 58.10         & 45.17 & \textcolor{ForestGreen}{12.93$\uparrow$} \\
        Chatglm2-6B    & 47.31    & 53.00     & 50.43     & 51.16        & 49.66 & \textcolor{ForestGreen}{1.50$\uparrow$}  \\
        Interlm2-7B    & 53.23    & 70.28  & 72.17     & 66.26        & 57.28 & \textcolor{ForestGreen}{8.98$\uparrow$} \\
        Llama2-7B      & 42.47    & 51.61  & 53.91     & 49.66        & 41.22 & \textcolor{ForestGreen}{8.44$\uparrow$} \\
        Vicuna-7b      & 45.16    & 54.15  & 52.17     & 51.56        & 51.43 & \textcolor{ForestGreen}{0.13$\uparrow$} \\ 
        \bottomrule
    \end{tabular}
    \caption{Direct guidance with schema knowledge on instance level CRR task. \textsc{w.t.S} and \textsc{w.o.S} stands for average performances with and without event knowledge guidance. $\Delta$ is the difference between them.}
    \label{tab:crr direct}
\end{table*}

\subsection{To what extent do LLMs have the event schema knowledge?}
In the previous section, we acknowledge that LLMs can complete event reasoning to some extent. However, whether they are endowed with event schema knowledge remains unknown. In this part, we mainly explore to what extent LLMs have the event schema knowledge, i.e. of the schema level.

\paragraph{LLMs have event schema knowledge.}
We evaluate CEC and CRR on the schema level. The results are shown in Figure~\ref{fig:main} and detailed results in the Appendix, and the average scores are reported in Table~\ref{tab:avg}. We find LLMs already have event schema knowledge and can complete both CEC and CRR tasks at the schema level to some extent. However, in Table~\ref{tab:avg}, we observe that S-CEC lags I-CEC, suggesting that LLMs are more adept at reasoning at the instance level.

\paragraph{Event schema knowledge increases falling behind reasoning at the instance level.}
We probe how event schema knowledge increases with the development of LLMs. We depict CEC and CRR performance comparisons of LLMs on instance- and schema-level in Figure~\ref{fig:S-I}. When the models initially can reason about events, they also have event schema knowledge. At this time, models can perform comparatively or even better in schema-level event reasoning. With the development, models perform instance-level reasoning on par with schema-level. It indicates that the accumulation of event schema knowledge falls behind the reasoning at the instance level. This finding demonstrates that enhancing event schema knowledge may further improve these abilities to obtain better general LLMs.

\subsection{Are LLMs aligned with humans in the aspect of leveraging event schema knowledge}
In this section, we investigate how LLMs leverage event schema knowledge to complete event reasoning. We first provide the instance-level question for the models and then ask them to generate the required event schema knowledge to solve the task. Then we evaluate the accuracy of the generated event schema knowledge.

Since we have the ground truth event schema knowledge for each question, the only challenge is to guide the LLMs to generate in a similar format for calculating accuracy. The instruction of our prompt first asks LLMs to generate the event types of each instance event in data. Based on the event types, it requires the LLMs to further generate relation triplets needed for the question. 

However, we find the LLMs would generate event types of different words but correct contents. To mitigate this problem, we prepare a list of candidate event types for each data to make it a classification setting. To keep the task difficult, we first conduct KMeans clustering on all event types in our dataset\footnote{We use \texttt{all-mpnet-base-v2} for encoding.}. We obtain 1000 clusters. For each data, we assign 20 random candidates in total including the correct ones. The negative event types are chosen from different clusters. 

After the generation, we calculate the accuracy of event types and F1-scores of relation triplets respectively comparing with the human-labeled event schema. We regard a correct triplet if all the head and tail event types and the inter-relation align with the human labels. We show detailed examples in the Appendix.

The results are in Table~\ref{tab:align}. We find only GPT4 and GPT4o can generate correct event types. GPT4 excels may be attributed to 1) its better alignment. 2) The dataset is originally generated by GPT4. Therefore in this part, we mainly evaluate other models rather than GPT4 series. However, we find rest models all fail to generate corresponding schema knowledge. For relation triplet generation, even GPT4 can not output proper event schemas. It significantly suggests that LLMs may not leverage event schema knowledge as humans when solving event reasoning tasks. Alignment of using such knowledge could further improve the performances.

\subsection{Can LLMs perform better event reasoning with explicit guidance of leveraging event schema knowledge?}
In the previous section, we find LLMs may not leverage event schema knowledge as human does. It raises an interesting question how well LLMs perform if we guide them to explicitly use such knowledge?
In this section, we probe this question.

We conduct experiments in which we directly add the schema event of each instance event in the question into the prompt. We also add relations of these schema events. 

We demonstrate the performances of this guidance in Table~\ref{tab:cec direct} and ~\ref{tab:crr direct}. We find incorporating event schema knowledge significantly improves event reasoning. It shows great potential to solve event reasoning with the fusion of event schema knowledge. These results provide important insights that event schema knowledge could be used as memory to improve solving practical and domain-specific problems.
In this view, constructing and retrieving proper event schema knowledge is another challenge. We leave them to future works.





\section{Related Work}
\noindent\paragraph{Event Reasoning}
\citet{du2022care} aims to select the accurate cause or effect event from candidates. \citet{zhou2019going} serves as a dataset for event temporal reasoning. Current works present a scenario of incorporating counterfactual reasoning~\cite{qin2019counterfactual, qin2020back}.
In addition to single-event relation reasoning, existing works also reason events according to diversified event relations~\cite{poria2021recognizing, han2021ester, yang2022towards}. \citet{tao2023unievent} further unifies datasets of several event-inter relations to transfer event relational knowledge to unseen tasks. 

Predicting events necessitates the model to anticipate forthcoming occurrences grounded in the present context ~\cite{zhao2021event}. \citet{mostafazadeh2016corpus} employs a multiple-choice framework to predict future events by encompassing a diverse range of common-sense connections among events. \citet{guan2019story} establish a dataset oriented towards capturing event logic, enabling the generative prediction of future incidents.

\noindent\paragraph{Evaluations for LLMs}
Evaluating the capacities of LLMs is the foundation of using and improving them. 
One group of research evaluates the general abilities of LLMs~\cite{hendrycks2020measuring, zheng2023judging, zhong2023agieval, bang2023multitask}
Besides, existing works evaluate LLMs in specific tasks~\cite{bang2023multitask, bian2023chatgpt,gao2023exploring,wei2023zero,li2024leveraging}.
Related to event reasoning, \citet{yuan2023zero} evaluated the ability to solve event relation extraction. \citet{tao2023eveval} present the Event Semantic Processing including the event understanding, reasoning, and prediction of event semantics. \citet{chan2023chatgpt} investigates relation reasoning between sentences. Compared with them, we are the first to introduce the evaluation for both schema- and instance-level event reasoning. Moreover, we comprehensively evaluate the performances of various relations and reasoning paradigms.


\section{Conclusion}
In this paper, we evaluate the event reasoning of LLMs. We introduce a novel benchmark \ees which features both levels of schema and instance. It evaluates event schema knowledge and reasoning abilities. Besides, \ees can be used to comprehensively evaluate the event reasoning in various relations and reasoning paradigms. We conduct extensive experiments on \ee. We obtain many insights such as: 1) LLMs have the abilities of event reasoning, but are far from satisfactory and are unbalanced in different relations and reasoning paradigms. 2) LLMs have a comprehension of event schema knowledge.  3) LLMs are not aligned with human to leaverage event schema knowledge in event reasoning. 4) Based on the findings, we guide the LLMs to utilize event schema knowledge. With our guidance, LLMs can perform better event reasoning.


\bibliography{main}
\newpage
\section*{Appendix}

\subsection*{Detailed Results}
We show detailed results in Table~\ref{detail_s_cec}~(S-CEC), Table~\ref{detail_i_cec}~(I-CEC), Table~\ref{detail_s_crr}~(S-CRR), and Table~\ref{detail_s_crr}~(I-CRR).

\begin{table*}
\centering
\footnotesize
\setlength\tabcolsep{3pt}\begin{tabular}{lcccccccccc}
\toprule
Model & \texttt{Before} & \texttt{After} & \texttt{Cause} & \texttt{Result} & \texttt{isSubevent} & \texttt{hasSubevent} & \texttt{Temporal} & \texttt{Causal} & \texttt{Hierarchy} & All \\
\midrule
GPT4o & 66.67 & 53.57 & 68.52 & 71.98 & 76.67 & 82.35 & 57.72 & 71.33 & 77.92 & 68.93 \\
GPT4 & 64.1 & 53.57 & 66.67 & 72.84 & 71.67 & 76.47 & 56.91 & 71.68 & 72.73 & 68.11 \\
GPT3.5 & 46.15 & 48.81 & 74.07 & 72.41 & 65.00 & 70.59 & 47.97 & 72.73 & 66.23 & 65.43 \\
Mistral-7B & 38.46 & 47.62 & 44.44 & 57.33 & 56.67 & 52.94 & 44.72 & 54.90 & 55.84 & 52.47 \\
Qwen2-7B & 56.41 & 50.00 & 57.41 & 68.53 & 63.33 & 58.82 & 52.03 & 66.43 & 62.34 & 62.14 \\
Baichuan2-7B & 51.28 & 48.81 & 53.7 & 55.17 & 48.33 & 58.82 & 49.59 & 54.90 & 50.65 & 52.88 \\
Orca2-7B & 53.85 & 42.86 & 64.81 & 67.67 & 55.00 & 52.94 & 46.34 & 67.13 & 54.55 & 59.88 \\
Chatglm2-6B & 38.46 & 45.24 & 57.41 & 61.21 & 55.00 & 70.59 & 43.09 & 60.49 & 58.44 & 55.76 \\
Interlm2-7B & 53.85 & 52.38 & 72.22 & 69.83 & 71.67 & 64.71 & 52.85 & 70.28 & 70.13 & 65.84 \\
Llama2-7B & 33.33 & 36.90 & 50.00 & 48.71 & 41.67 & 58.82 & 35.77 & 48.95 & 45.45 & 45.06 \\
Vicuna-7b & 15.38 & 28.57 & 16.67 & 28.02 & 33.33 & 11.76 & 24.39 & 25.87 & 28.57 & 25.93 \\
\bottomrule
\end{tabular}
\caption{Detailed results on S-CEC.}
\label{detail_s_cec}
\end{table*}

\begin{table*}
    \centering
    \footnotesize
\setlength\tabcolsep{3pt}\begin{tabular}{lcccccccccc}
        \toprule
Model & \texttt{Before} & \texttt{After} & \texttt{Cause} & \texttt{Result} & \texttt{isSubevent} & \texttt{hasSubevent} & \texttt{Temporal} & \texttt{Causal} & \texttt{Hierarchy} & All \\
        \midrule
        GPT4o         & 60.00     & 52.38 & 69.09 & 70.64  & 75.00         & 58.82       & 54.84    & 70.34  & 71.43     & 66.60 \\
        GPT4          & 70.00     & 58.33 & 69.09 & 69.36  & 76.67      & 70.59       & 62.1     & 69.31  & 75.32     & 68.43 \\
        GPT3.5        & 50.00     & 44.05 & 74.55 & 71.06  & 73.33      & 52.94       & 45.97    & 71.72  & 68.83     & 64.77 \\
        Mistral-7B    & 45.00     & 35.71 & 63.64 & 62.98  & 45.00         & 47.06       & 38.71    & 63.10   & 45.45     & 54.18 \\
        Qwen2-7B      & 55.00     & 52.38 & 76.36 & 72.34  & 50.00         & 29.41       & 53.23    & 73.10   & 45.45     & 63.75 \\
        Baichuan2-7B  & 22.5   & 26.19 & 43.64 & 29.36  & 30.00         & 29.41       & 25.00       & 32.07  & 29.87     & 29.94 \\
        Orca2-7B      & 55.00     & 47.62 & 56.36 & 67.66  & 55.00         & 58.82       & 50.00       & 65.52  & 55.84     & 60.08 \\
        Chatglm2-6B   & 32.50   & 9.52  & 50.91 & 33.19  & 33.33      & 29.41       & 16.94    & 36.55  & 32.47     & 30.96 \\
        Interlm2-7B   & 57.50   & 40.48 & 67.27 & 68.09  & 68.33      & 58.82       & 45.97    & 67.93  & 66.23     & 62.12 \\
        Llama2-7B     & 32.5   & 28.57 & 45.45 & 25.53  & 33.33      & 23.53       & 29.84    & 29.31  & 31.17     & 29.74 \\
        Vicuna-7b     & 22.5   & 29.76 & 20.00    & 28.94  & 25.00         & 29.41       & 27.42    & 27.24  & 25.97     & 27.09 \\
        \bottomrule
    \end{tabular}
\caption{Detailed results on I-CEC.}
\label{detail_i_cec}
\end{table*}

\begin{table*}
\centering
\footnotesize
\setlength\tabcolsep{10pt}\begin{tabular}{lcccc}
\toprule
Model & \texttt{Temporal} & \texttt{Causal} & \texttt{Hierarchy} & All \\
\midrule
GPT4o & 51.89 & 66.98 & 60.00 & 62.05 \\
GPT4 & 61.08 & 66.28 & 53.91 & 63.01 \\
GPT3.5 & 46.49 & 58.60 & 52.17 & 54.52 \\
Mistral-7B & 46.49 & 55.58 & 45.22 & 51.64 \\
Qwen2-7B & 41.62 & 58.14 & 46.09 & 52.05 \\
Baichuan2-7B & 49.73 & 53.49 & 47.83 & 51.64 \\
Orca2-7B & 33.51 & 52.79 & 41.74 & 46.16 \\
Chatglm2-6B & 50.27 & 56.51 & 40.87 & 52.47 \\
Interlm2-7B & 45.41 & 51.63 & 42.61 & 48.63 \\
Llama2-7B & 35.14 & 51.40 & 46.96 & 46.58 \\
Vicuna-7B & 47.57 & 53.95 & 52.17 & 52.05 \\
\bottomrule
\end{tabular}
\caption{Detailed results on S-CRR.}
\label{detail_s_crr}
\end{table*}

\begin{table*}
    \centering
    \footnotesize
\setlength\tabcolsep{10pt}\begin{tabular}{lcccc}
        \toprule
        Model & \texttt{Temporal} & \texttt{Causal} & \texttt{Hierarchy} & All \\
        \midrule
        GPT4o & 48.92 & 67.74 & 61.74 & 62.04 \\
        GPT4 & 52.15 & 67.05 & 70.43 & 63.81 \\
        GPT3.5 & 36.56 & 46.54 & 46.09 & 43.95 \\
        Mistral-7B & 43.55 & 62.44 & 46.96 & 55.24 \\
        Qwen2-7B & 36.56 & 53.69 & 42.61 & 47.62 \\
        Baichuan2-7B & 36.02 & 51.38 & 37.39 & 45.31 \\
        Orca2-7B & 41.94 & 49.77 & 33.04 & 45.17 \\
        Chatglm2-6B & 46.77 & 54.15 & 37.39 & 49.66 \\
        Interlm2-7B & 43.55 & 61.29 & 64.35 & 57.28 \\
        Llama2-7B & 35.48 & 43.55 & 41.74 & 41.22 \\
        Vicuna-7b & 44.62 & 54.15 & 52.17 & 51.43 \\
        \bottomrule
    \end{tabular}
\caption{Detailed results on I-CRR.}
\label{detail_i_crr}
\end{table*}

\begin{table*}[h]
\centering
\footnotesize
\setlength\tabcolsep{1pt}
\begin{tabular}{lll}

\toprule
\toprule

 Model & Version & URL \\
\midrule
GPT4o &  gpt-4o-2024-05-13 &  \\
GPT4~\cite{achiam2023gpt} &  gpt-4-0125-preview &  \\
GPT3.5 &  gpt-3.5-turbo-1106 &  \\
Mistral-7B~\cite{jiang2023mistral} & mistralai/Mistral-7B-Instruct-v0.2 & \href{https://huggingface.co/mistralai/Mistral-7B-Instruct-v0.2}{https://huggingface.co/mistralai/Mistral-7B-Instruct-v0.2} \\
Qwen2-7B~\cite{yang2024qwen2} & Qwen/Qwen2-7B-Instruct & \href{https://huggingface.co/Qwen/Qwen2-7B-Instruct}{https://huggingface.co/Qwen/Qwen2-7B-Instruct}\\
Baichuan2-7B~\cite{yang2023baichuan} & baichuan-inc/Baichuan2-7B-Chat & \href{https://huggingface.co/baichuan-inc/Baichuan2-7B-Chat}{https://huggingface.co/baichuan-inc/Baichuan2-7B-Chat} \\
Orca2-7B~\cite{mitra2023orca} & microsoft/Orca-2-7b & \href{https://huggingface.co/microsoft/Orca-2-7b}{https://huggingface.co/microsoft/Orca-2-7b} \\
Chatglm2-7B~\cite{glm2024chatglm} & THUDM/chatglm2-6b & \href{https://huggingface.co/THUDM/chatglm2-6b}{https://huggingface.co/THUDM/chatglm2-6b} \\
Internlm2-7B~\cite{cai2024internlm2} & internlm/internlm2\_5-7b-chat & \href{https://huggingface.co/internlm/internlm2\_5-7b-chat}{https://huggingface.co/internlm/internlm2\_5-7b-chat} \\
Llama2-7B~\cite{touvron2023llama} & NousResearch/Llama-2-7b-chat-hf & \href{https://huggingface.co/NousResearch/Llama-2-7b-chat-hf}{https://huggingface.co/NousResearch/Llama-2-7b-chat-hf} \\
Vicuna-7B~\cite{vicuna2023} & lmsys/vicuna-7b-v1.3 & \href{https://huggingface.co/lmsys/vicuna-7b-v1.3}{https://huggingface.co/lmsys/vicuna-7b-v1.3} \\

\bottomrule
\end{tabular}
\caption{Evaluated model versions and urls.}
\label{tab: model detail}
\end{table*}

\subsection*{Annotation Process}

\begin{figure*}[!tb]
    \centering
    \includegraphics[width=2.1\columnwidth]{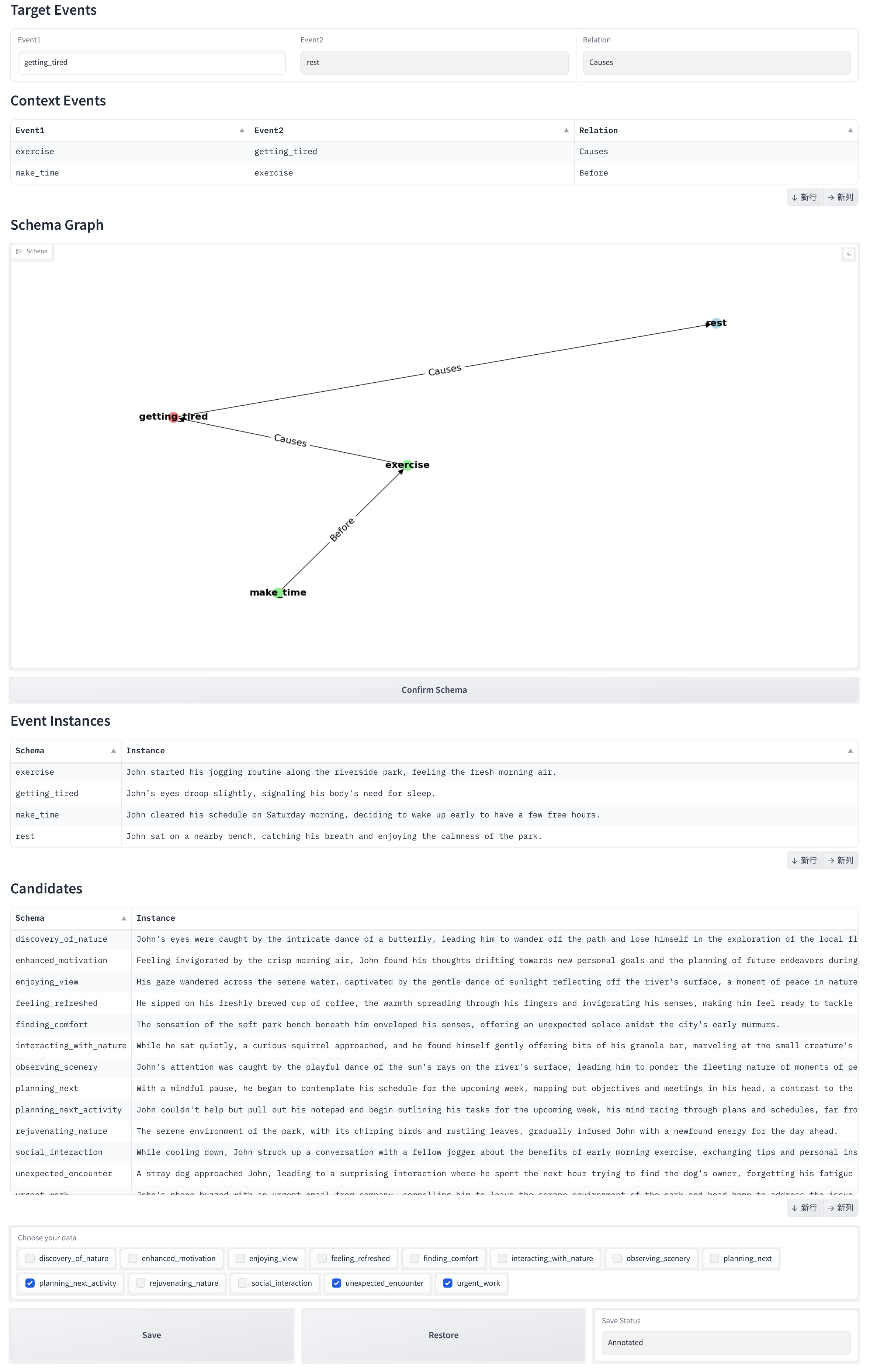}
    \caption{Annotation platform. This Figure shows the process of annotating one data.}
    \label{fig:platform}
\end{figure*}

In this section, we describe the annotation of the third stage in detail. We illustrate the annotation platform in Figure~\ref{fig:platform}. After logining in and selecting assigned data (i.e. the prompting graphs and prepared candidate events), to annotate this data, the annotator should undergo three steps:
\begin{itemize}[topsep=0pt]
    \setlength{\itemsep}{0pt}
    \setlength{\parskip}{0pt}
\setlength{\leftmargin}{-1pt}
\item[1)] \textbf{Modify the schema graph} As shown in Figure~\ref{fig:platform}, Target Events are the queried event, answer event, and their relation. Context Events are other events in schema graph. The Event Schema Graph shows the queried event in red, answer event in blue, and other context event in green. The first step is to rewrite the events and relations in in schema graph to make them strictly valid. 

\item[2)] \textbf{Modify the instance graph} The Event Instances are the instance events of the events in the schema graph. The second step is to rewrite the instance events to make the instance graph valid and is relationally consistent with the schema graph. 

\item[3)] \textbf{Modify candidate events} The Candidates are the events in both schema and instance levels. They are the candidates of the answer event in the final CEC question. The last step is to modify and select candidates as negative choices considering the answer event where each candidate event consists of a schema and an instance event.

\end{itemize}

\subsection*{Model Details}
We list all model details in Table~\ref{tab: model detail}.

\clearpage

\begin{figure*}[htbp]
\begin{tcolorbox}[title=Prompt of schema-level CEC, label=fig7, enhanced,colback=gray!5!white,colframe=gray!75!black,drop fuzzy shadow southwest,fontupper=\small]
\textbf{\#\#\# Instructions:} \\
Answer the question by selecting A, B, C, D. \\
\\
\textbf{\#\#\# Context:} \\
"study" is a subevent of "analyse". "analyse" is after
"think". "pass$\textunderscore$class" is after "study". \\

\textbf{\#\#\# Question:} \\
Which event has the subevent of "think"? Choices: \\
A. research \\
B. attend$\textunderscore$conference \\
C. plan$\textunderscore$project \\
D. talk$\textunderscore$to The answer is \\
\\
The answer is  
\end{tcolorbox}
\label{prompt_s-cec}
\end{figure*}

\begin{figure*}[h]
\begin{tcolorbox}[title= Prompt of instance-level CEC, enhanced,colback=gray!5!white,colframe=gray!75!black,drop fuzzy shadow southwest,fontupper=\small]
    \textbf{\#\#\# Instructions:} \\
Answer the question by selecting A, B, C, D. Note that all events appearing in
"Context", "Question", and "Choices" refer to the specific events described in
"Instances". \\
\\
\textbf{\#\#\# Instances:}  \\
event9:  \\
In anticipation of the challenging final exams, she had dedicated countless evenings in the library, poring over textbooks and academic papers on climate change.  \\
event65:  \\
These insights were further refined through discussions with her mentor, Professor Ramirez, who provided valuable feedback and perspectives. \\
event30: \\
One pivotal moment was when she deciphered the complex data on global
warming trends, developing a comprehensive presentation. \\
event99: \\
After months of preparation, Alice finally received her diploma in Environmental Science from the university. \\
event86: \\
This breakthrough came after she spent a weekend in solitude at a cabin in the
woods, reflecting on the interconnectedness of natural systems. \\
event32: \\
After her deep reflections, she crafted an ambitious project plan aiming to initiate a community-driven reforestation program, outlining steps for local engagement and environmental restoration efforts. \\
event5: \\
Alice attended an international conference on sustainable development, where she presented her findings on the effectiveness of renewable energy sources in
reducing carbon emissions. \\
event90: \\
She conducted an in-depth analysis of historical environmental policy reforms to understand their impact on current climate advocacy strategies. \\
\\
\textbf{\#\#\# Context:} \\
"event9" is a subevent of "event30". "event30" is after "event86". "event99" is after "event9".  \\
\\
\textbf{\#\#\# Question:} \\
Which event has the subevent of "event86"? \\
Choices: \\
A. event32 \\
B. event65 \\
C. event90 \\
D. event5 \\
\\
The answer is 
\end{tcolorbox}
\label{prompt_i-cec}
\end{figure*}


\begin{figure*}[htbp]
\begin{tcolorbox}[title=Prompt of schema-level CRR, enhanced,colback=gray!5!white,colframe=gray!75!black,drop fuzzy shadow southwest,fontupper=\small]
    \textbf{\#\#\# Instructions:} \\
Answer the question by selecting A, B, or C. \\
\\
\textbf{\#\#\# Context:} \\
"fall" is a subevent of "exercise". "miss" causes "fall". "miss" causes "exercise". \\
\\
\textbf{\#\#\# Question:} \\
Which is the causal relationship between "fall" and "lack$\textunderscore$energy"? \\
Choices: \\
A. "fall" causes "lack$\textunderscore$energy".  \\
B. "fall" is result of "lack$\textunderscore$energy".  \\
C. There is no obvious causal relationship between "fall" and "lack$\textunderscore$energy".  \\
\\
The answer is
\end{tcolorbox}
\label{prompt_s-crr}
\end{figure*}


\begin{figure*}[htbp]
\begin{tcolorbox}[title= Prompt of instance-level CRR, enhanced,colback=gray!5!white,colframe=gray!75!black,drop fuzzy shadow southwest,fontupper=\small]
    \textbf{\#\#\# Instructions:} \\
Answer the question by selecting A, B or C. Note that all events appearing in "Context", "Question", and "Choices" refer to the specific events described in "Instances". \\
\\
\textbf{\#\#\# Instances:} \\
event66: \\
Sitting in the second row, the jurors leaned forward, focusing intently on every word spoken by the witness, understanding the gravity of the details being shared. \\
event88: \\
The court case of John Doe for alleged embezzlement commenced on a rainy Monday morning at the downtown courthouse. \\
event90: \\
During the proceedings, a key witness was called to the stand to provide a detailed account of the financial transactions in question. \\
\\
\textbf{\#\#\# Context:}  \\
"event88" causes "event90". \\
\\
\textbf{\#\#\# Question:} \\
Which is the subordinate relationship between "event90" and "event66"? \\
Choices: \\
A. "event66" is subevent of "event90". \\
B. "event90" is subevent of "event66". \\
C. There is no obvious subordinate relationship between "event90" and "event66". \\
\\
The answer is
\end{tcolorbox}
\label{prompt_i-crr}
\end{figure*}


\begin{figure*}[htbp]
\begin{tcolorbox}[title=A example of detailed Prompt and ground truth for alignment evaluation on CEC, enhanced,colback=gray!5!white,colframe=gray!75!black,drop fuzzy shadow southwest,fontupper=\scriptsize]
    \textbf{\#\#\# Instructions:} \\
In a scenario explained by the "Context", the "Question" ask about selecting one relationship between two events, from all possible relationships provided by the "Choices", and the "Instances" explain all
the events in detail with a few sentences. Event semantic knowledge refers to the abstract event types to which specific events belong, and the relationships between these abstract event types. Please
output the event semantic knowledge used in solving the following problem. Note that all possible abstract event categories in the "Schema", and the relationships between abstract events include
HasSubevent, IsSubevent, Before, After, Causes, and IsResult. For the tuple [event0, relation, event1], HasSubevent indicates that event1 is a subevent of event0, IsSubevent indicates that event0 is a
subevent of event1, Before indicates that event0 occurs before event1, After indicates that event0 occurs after event1, Causes indicates that event0 causes event1, and IsResult indicates that event0 is the
result of event1. Output in JSON format, don't generate other sentences. \\
\\
\textbf{\#\#\# Requirements:} \\
Abstract event types can only be chosen from "Schema", and the relationships of abstract event types can only be selected from HasSubevent, IsSubevent, Before, After, Causes, and IsResult. Follow the
format in examples, output in JSONL format. The key "event type" should correspond to a value that is a dictionary with events as keys and their abstracted categories as values. The key "event relation"
should correspond to a value that is a list of tuples [event0, relation, event1]. The relationships between events include HasSubevent, IsSubevent, Before, After, Causes, and IsResult. \\
\\
\textbf{\#\#\# Schema:} \\
artistic$\textunderscore$innovation, relocate, pass$\textunderscore$class, experience$\textunderscore$emotional$\textunderscore$distress, seek$\textunderscore$guidance, think, drinking$\textunderscore$coffee, answer, attend$\textunderscore$conference, tell$\textunderscore$lies, feeling$\textunderscore$homesick, bored, review$\textunderscore$notes, research, talk$\textunderscore$to, study, plan$\textunderscore$project, getting$\textunderscore$exercise, analyse, gaining$\textunderscore$recognition
\\
Here are some examples: \\
\\
\textbf{\#\#\# Instances:} \\
event53: \\
This revelation ultimately prompted an individual in the courtroom audience to discretely exit the room and later that evening, the same individual, driven by fear of exposure, went on to commit a fatal
assault against a witness who could connect him to the crime. \\
event67: \\
While Mr. Smith was providing his account, he mentioned a key detail that was previously overlooked—a unique tattoo that he glimpsed on the perpetrator's arm. \\
event91: \\
Two weeks later, during the heated court proceedings at the downtown courthouse, the homeowner, Mr. Smith, was called to testify before the jury regarding the night of the incident. \\
event64: \\
In a quiet suburban neighborhood, a burglary occurred at the Smith residence, where an unknown assailant broke in and stole valuable heirlooms late at night. \\
\\
\textbf{\#\#\# Context:}
"event64" is before "event91". "event67" is a subevent of "event91". \\
\\
\textbf{\#\#\# Question:} \\
Which is the causal relationship between "event67" and "event53"? \\
Choices: \\
A. "event67" causes "event53". \\
B. "event67" is result of "event53". \\
C. There is no obvious causal relationship between "event67" and "event53". \\
\\
Event type and event relation: \\
{"event$\textunderscore$type": {"event67": "talk", "event53": "kill", "event64": "commit$\textunderscore$crime", "event91": "take$\textunderscore$stand"}, "event$\textunderscore$relation": [["commit$\textunderscore$crime", "Before", "take$\textunderscore$stand"], ["take$\textunderscore$stand", "HasSubevent", "talk"],
["talk", "Causes", "kill"]]} \\
\\
Now, based on the above, please output the event semantic knowledge used in solving the following problem. \\
\\
\textbf{\#\#\# Instances:} \\
event66: \\
Sitting in the second row, the jurors leaned forward, focusing intently on every word spoken by the witness, understanding the gravity of the details being shared. \\
event88: \\
The court case of John Doe for alleged embezzlement commenced on a rainy Monday morning at the downtown courthouse. \\
event90: \\
During the proceedings, a key witness was called to the stand to provide a detailed account of the financial transactions in question. \\
\\
\textbf{\#\#\# Context:} \\
"event88" causes "event90". \\
\textbf{\#\#\# Question:} \\
Which is the subordinate relationship between "event90" and "event66"? \\
Choices: \\
A. "event66" is subevent of "event90". \\
B. "event90" is subevent of "event66". \\
C. There is no obvious subordinate relationship between "event90" and "event66".\\
\\
Event type and event relation:
\\
\\
\textbf{\#\#\# Ground Truth:} \\
{"event$\textunderscore$type": {"event90": "hear$\textunderscore$testimony", "event88": "trial", "event66": "paying$\textunderscore$attention"}, "event$\textunderscore$relation": [["trial", "Causes", "hear$\textunderscore$testimony"], ["hear$\textunderscore$testimony", "HasSubevent", "paying$\textunderscore$attention"]]} 
\end{tcolorbox}
\label{rq3_detailed_cec}
\end{figure*}



\begin{figure*}[htbp]
\begin{tcolorbox}[title= A example of detailed Prompt and ground truth for alignment evaluation on CRR, enhanced,colback=gray!5!white,colframe=gray!75!black,drop fuzzy shadow southwest,fontupper=\scriptsize]
    \textbf{\#\#\# Instructions:} \\
In a scenario explained by the "Context", the "Question" ask about selecting one relationship between two events, from all possible relationships provided by the "Choices", and the "Instances" explain all
the events in detail with a few sentences. Event semantic knowledge refers to the abstract event types to which specific events belong, and the relationships between these abstract event types. Please
output the event semantic knowledge used in solving the following problem. Note that all possible abstract event categories in the "Schema", and the relationships between abstract events include
HasSubevent, IsSubevent, Before, After, Causes, and IsResult. For the tuple [event0, relation, event1], HasSubevent indicates that event1 is a subevent of event0, IsSubevent indicates that event0 is a
subevent of event1, Before indicates that event0 occurs before event1, After indicates that event0 occurs after event1, Causes indicates that event0 causes event1, and IsResult indicates that event0 is the
result of event1. \\
\\
\textbf{\#\#\# Requirements:} \\
Abstract event types can only be chosen from "Schema", and the relationships of abstract event types can only be selected from HasSubevent, IsSubevent, Before, After, Causes, and IsResult. Follow the
format in examples, output in JSONL format. The key "event type" should correspond to a value that is a dictionary with events as keys and their abstracted categories as values. The key "event relation"
should correspond to a value that is a list of tuples [event0, relation, event1]. The relationships between events include HasSubevent, IsSubevent, Before, After, Causes, and IsResult.\\
\\
\textbf{\#\#\# Schema:} \\
injury, write$\textunderscore$story, review$\textunderscore$appeal, competing, improving$\textunderscore$skill, delivering$\textunderscore$verdict, experience$\textunderscore$illness, competing$\textunderscore$against, discuss$\textunderscore$results, engage$\textunderscore$physical$\textunderscore$activity, trial, plan, gaining$\textunderscore$weight, hear$\textunderscore$testimony,
conduct$\textunderscore$research, develop$\textunderscore$immunity, decorate$\textunderscore$venue, organize$\textunderscore$thoughts, educate$\textunderscore$child, paying$\textunderscore$attention \\
\\
Here are some examples: \\
\\
\textbf{\#\#\# Instances:} \\
event53: \\
This revelation ultimately prompted an individual in the courtroom audience to discretely exit the room and later that evening, the same individual, driven by fear of exposure, went on to commit a fatal
assault against a witness who could connect him to the crime. \\
event67:\\
While Mr. Smith was providing his account, he mentioned a key detail that was previously overlooked—a unique tattoo that he glimpsed on the perpetrator's arm. \\
event91: \\
Two weeks later, during the heated court proceedings at the downtown courthouse, the homeowner, Mr.Smith, was called to testify before the jury regarding the night of the incident. \\
event64: \\
In a quiet suburban neighborhood, a burglary occurred at the Smith residence, where an unknown assailant broke in and stole valuable heirlooms late at night. \\
\\
\textbf{\#\#\# Context:} \\
"event64" is before "event91". "event67" is a subevent of "event91". \\
\\
\textbf{\#\#\# Question:} \\
Which is the causal relationship between "event67" and "event53"? \\
Choices: \\
A. "event67" causes "event53".\\
B. "event67" is result of "event53".\\
C. There is no obvious causal relationship between "event67" and "event53".\\
Event type and event relation:\\
{"event$\textunderscore$type": {"event67": "talk", "event53": "kill", "event64": "commit$\textunderscore$crime", "event91": "take$\textunderscore$stand"}, "event$\textunderscore$relation": [["commit$\textunderscore$crime", "Before", "take$\textunderscore$stand"], ["take$\textunderscore$stand", "HasSubevent", "talk"],
["talk", "Causes", "kill"]]} \\
\\
Now, based on the above, please output the event semantic knowledge used in solving the following problem. \\
\\
\textbf{\#\#\# Instances:} \\
event66:\\
Sitting in the second row, the jurors leaned forward, focusing intently on every word spoken by the witness, understanding the gravity of the details being shared.\\
event88:\\
The court case of John Doe for alleged embezzlement commenced on a rainy Monday morning at the downtown courthouse.\\
event90:\\
During the proceedings, a key witness was called to the stand to provide a detailed account of the financial transactions in question.\\
\\
\textbf{\#\#\# Context:} \\
"event88" causes "event90". \\
\\
\textbf{\#\#\# Question:} \\
Which is the subordinate relationship between "event90" and "event66"? \\
Choices: \\
A. "event66" is subevent of "event90". \\
B. "event90" is subevent of "event66". \\
C. There is no obvious subordinate relationship between "event90" and "event66". \\
\\
Event type and event relation: 
\\
\\
\textbf{\#\#\# Ground Truth:} \\
{"event$\textunderscore$type": {"event90": "hear$\textunderscore$testimony", "event88": "trial", "event66": "paying$\textunderscore$attention"}, "event$\textunderscore$relation": [["trial", "Causes", "hear$\textunderscore$testimony"], ["hear$\textunderscore$testimony", "HasSubevent", "paying$\textunderscore$attention"]]}
\end{tcolorbox}
\label{rq3_detailed_crr}
\end{figure*}

\end{document}